\documentclass[conference]{IEEEtran}
\IEEEoverridecommandlockouts

\usepackage{cite}
\usepackage{amsmath,amssymb,amsfonts}
\usepackage{algorithm}
\usepackage{algorithmic}
\usepackage{graphicx}
\usepackage{textcomp}
\usepackage{xcolor}
\usepackage{subcaption}

\usepackage{colortbl}
\usepackage{dhucs} 
\usepackage{booktabs} 
\usepackage{multirow}
\usepackage{comment}
\usepackage{lineno}
\usepackage{color} 
\usepackage{setspace} 
\usepackage{makecell}
\usepackage{enumitem} 
\usepackage{relsize}
\usepackage{tabularx}
\usepackage{diagbox}
\usepackage{bm}
\usepackage{hhline}

\usepackage{tikz}
\usepackage{pgfplots}
\usepackage{xurl}
\usepackage[colorlinks=false, pdfborder={0 0 0}]{hyperref}

\usepackage[a4paper, total={184mm,243mm}]{geometry}
\def\BibTeX{{\rm B\kern-.05em{\sc i\kern-.025em b}\kern-.08em
    T\kern-.1667em\lower.7ex\hbox{E}\kern-.125emX}}

\usepackage[font=small,labelfont=small]{caption}

\usepackage{siunitx}
\makeatletter
\renewcommand\paragraph{%
  \@startsection{paragraph}{4}{0.8em}
    {1.0ex plus .2ex minus .2ex}
    {-0.5em}
    {\normalfont\normalsize}%
}
\makeatother

\newcommand\korean[1]{}

\newcommand\refFigure[1]{Fig.~\ref{#1}}
\newcommand\refTable[1]{Table~\ref{#1}}

\newcommand\refAlgo[1]{Algorithm~\ref{#1}}
\newcommand\refEqn[1]{(\ref{#1})}

\newcommand\blind[1]{XXXX}

\begin{document}

\title{\LARGE\relsize{+1} LoRA-Edge: Tensor-Train–Assisted LoRA for Practical CNN Fine-Tuning on Edge Devices}

\IEEEaftertitletext{\vspace{-2ex}}

\author{
Hyunseok Kwak$^{ \dag \sharp}$, Kyeongwon Lee$^{  \dag \sharp}$, Jae-Jin Lee$^{\ddag}$, and Woojoo Lee$^{ \dag  \ast}$\\
$^{ \dag}$School of Intelligent Semiconductor Engineering, Chung-Ang University, Seoul, Korea \\
$^{ \ddag}$AI Edge SoC Research Section, the Electronics and Telecommunications Research Institute, Deajeon, Korea \\

\thanks{
This paper has been accepted for publication at the Design, Automation and Test in Europe (\textit{DATE 2026}). 
This document represents the camera-ready version.

$^{\sharp}$ Hyunseok Kwak and Kyeongwon Lee contributed equally to this work.

$^{*}$Woojoo Lee is the corresponding author.}
}

\maketitle


\begin{abstract}
On-device fine-tuning of CNNs is essential to withstand domain shift in edge applications such as Human Activity Recognition (HAR), yet full fine-tuning is infeasible under strict memory, compute, and energy budgets.  
We present \textit{LoRA-Edge}, a parameter-efficient fine-tuning (PEFT) method that builds on Low-Rank Adaptation (LoRA) with tensor-train assistance. LoRA-Edge (i) applies Tensor-Train Singular Value Decomposition (TT-SVD) to \emph{pre-trained} convolutional layers, (ii) \emph{selectively} updates only the output-side core with zero-initialization to keep the auxiliary path inactive at the start, and (iii) \emph{fuses} the update back into dense kernels, leaving inference cost unchanged. 
This design preserves convolutional structure and reduces the number of trainable parameters by up to two orders of magnitude compared to full fine-tuning.
Across diverse HAR datasets and CNN backbones, LoRA-Edge achieves accuracy within \textbf{4.7\%} of full fine-tuning while updating at most \textbf{1.49\%} of parameters, consistently outperforming prior parameter-efficient baselines under similar budgets. On a Jetson Orin Nano, TT-SVD initialization and selective-core training yield \textbf{1.4--3.8}$\times$ faster convergence to target F1. LoRA-Edge thus makes structure-aligned, parameter-efficient on-device CNN adaptation practical for edge platforms.
\end{abstract}

\section{Introduction}
Edge devices continue to gain richer sensing stacks and higher on-device compute, accelerating the practical adoption of edge-AI. 
Among edge workloads, Human Activity Recognition (\textit{HAR}) stands out because it transforms heterogeneous signals from wearable and ambient sensors into actionable context for healthcare, fitness, and smart IoT systems. 
In-the-wild deployments, however, rarely match the training distribution: user-specific motion patterns, sensor placement and orientation, device-dependent sampling, and other data characteristics frequently drift, producing chronic domain shift and accuracy degradation~\cite{Zhao:THMS20, Khaked:UbiComp23, Kasim:IJCNN23, Hong:IMWUT24}. 
Maintaining accuracy therefore requires {on-device}, continual fine-tuning rather than one-shot offline training~\cite{Li:TMC23, Liu:ICASSP24, Kang:IMWUT24}.

Full fine-tuning is at odds with edge constraints. Updating all weights demands significant memory traffic, compute, and energy under strict latency and power budgets, making this approach impractical on typical edge SoCs~\cite{Park:Date23, Park:TCASI24, CHOI:AEJ25, Jeon:DAC25}. 
Parameter-Efficient Fine-Tuning (\textit{PEFT}) is consequently the most realistic path forward. 
Representative PEFT families include Adapter modules~\cite{Houlsby:PMLR19}, Low-Rank Adaptation (\textit{LoRA})~\cite{Hu:ICLR22}, Bias-Tuning~\cite{Ben-Zaken:ACL22}, and Batch Normalization (\textit{BN}) tuning. 
Bias-Tuning and BN-Tuning may seem appealing due to their extremely small trainable footprint and ease of application to Convolutional Neural Networks (\textit{CNN}s). 
Nevertheless, in many HAR scenarios they fail to recover sufficient accuracy, limiting their practical value.
Adapter-based PEFT improves adaptivity but inserts additional modules between backbone layers; those modules persist after training and increase inference-time operations and memory—undesirable in resource-constrained settings.


LoRA has become popular because its low-rank adapters are trained along an auxiliary path and then merged into the base weights post-training, leaving inference cost unchanged. 
However, LoRA was designed for linear layers (as in Large Language Models, \textit{LLM}s) and does not directly accommodate the multi-dimensional structure of CNN convolutional layers, which include output channels, input channels, and spatial dimensions. 
A straightforward workaround flattens a convolutional weight tensor into a matrix and applies LoRA to the reshaped weights (e.g., LoRA-C~\cite{Ding:arXiv24}). 
This flattening, however, ties the adapter rank to the kernel size (for example, causing trainable parameters to scale with the square of kernel size), which often forces much larger trainable adapters to reach competitive accuracy.
From an edge on-device learning perspective, the resulting parameter count and training burden remain too high, especially for HAR where CNNs are the dominant backbone family~\cite{Wang:IJCNN17, Chen:CSUR21, Park:NIPS24}. 
In short, directly adopting PEFT methods developed for LLMs into CNN-based models on edge devices is ineffective in practice.

A more structure-aligned alternative is to operate in the native tensor domain of convolutional layers using Tensor-Train Decomposition (\textit{TTD}). 
TTD factorizes a high-order tensor into a sequence of 3D \emph{cores}, with truncation ranks controlling the parameter budget. Recent PEFT methods (e.g., LoRETTA~\cite{Yang:naacl24} and TT-LoRA~\cite{Anjum:ICMLA24}) exploit tensor trains for linear layers in LLMs by inserting randomly initialized TT adapters along an auxiliary path. 
While this leverages multiway structure, random initialization discards the rich feature representations already encoded in pre-trained CNN convolutional layers and requires learning them from scratch. For CNN-based HAR, where fine-tuning must be frequent and lightweight, such cold-start behavior inflates convergence time and energy—directly undermining practicality on edge devices.

This work proposes \textbf{\textit{LoRA-Edge}}, a PEFT framework that aligns low-rank adaptation with the multi-dimensional structure of CNNs while retaining the operational advantages of LoRA for inference.
Instead of inserting a separate, randomly initialized adapter, LoRA-Edge applies Tensor-Train Singular Value Decomposition (\textit{TT-SVD})~\cite{Oseledets:SIAM11} directly to \emph{pre-trained} convolutional weights to obtain a tensor-train factorization. 
We then \emph{selectively train} only a small subset of TT cores while keeping the remaining cores fixed. 
The number of trainable parameters scales with both TT-rank and kernel mode sizes, and LoRA-Edge further reduces this budget by updating only selected cores. 
Because initialization derives from the base model, optimization starts from an informed point that preserves useful feature structure, enabling faster convergence compared to randomly initialized TT adapters. 
After fine-tuning, the updated cores are \emph{fused back} to dense kernels and replace the original weights; 
thus, the inference-time FLOPs and memory footprint are identical to the unadapted backbone—retaining the merge-and-run property that makes LoRA attractive, while resolving the dimensional mismatch for CNNs without flattening.

Our goal is a practical on-device fine-tuning method for CNN models that (i) preserves convolutional structure without matrix flattening, 
(ii) minimizes trainable parameters and SRAM/DRAM pressure, (iii) converges quickly from pre-trained features, and (iv) preserves inference cost after training. 
To demonstrate generality, we validate LoRA-Edge on diverse CNN backbones (CALANet, T-ResNet, MobileNet) and datasets (Opportunity, DSADS, RealWorld, RealDisp), which exhibit challenging domain shifts.

Our contributions are summarized as follows:
\begin{enumerate}[leftmargin=*]
\item \textbf{Structure-preserving PEFT for CNNs.}
We directly decompose {multi-dimensional structure} with TT-SVD and perform adaptation in the tensor-train domain, eliminating the matrix-flattening mismatch inherent to LoRA for convolutional layers.
This preserves pre-trained structure and aligns parameterization with channel and spatial modes.

\item \textbf{Selective core training for edge budgets.}
We introduce a strategy that trains only the most impactful TT cores under explicit rank constraints, sharply reducing trainable parameters and memory traffic while accelerating convergence. Updated cores are merged post hoc, so inference-time compute and memory remain unchanged.

\item \textbf{Comprehensive validation.}
Across Opportunity, DSADS, RealWorld, and RealDisp with CALANet, T-ResNet, and MobileNet backbones, LoRA-Edge performs on-device adaptation using at most $1.49\%$ of the parameters updated by full fine-tuning, while limiting the accuracy gap to no more than 4.7\% relative to full fine-tuning—demonstrating efficiency and practicality under strict edge budgets.
\end{enumerate}

Overall, LoRA-Edge retains the operational simplicity of merge-after-training adaptation, but replaces matrix-based low-rank updates with a structure-aligned, TT-SVD–initialized pathway tailored to CNNs. 
By exploiting information already present in pre-trained convolutional layers and updating only selected TT cores, LoRA-Edge achieves fast, low-overhead on-device fine-tuning without altering inference cost. 
In doing so, this work establishes that on-device learning for CNNs can move from a theoretical possibility to a practical reality on resource-limited edge devices.

\section{PEFT for CNNs: Foundations and Related Work}

\subsection{LoRA: From Linear Layers to Convolutions}
$M$ and $N$ denote the output and input dimensions of a linear layer, respectively; 
$r$ is the LoRA rank with $r \ll \min(M,N)$; 
and $\#(\cdot)$ indicates the number of parameters. 
LoRA fine-tunes a model by freezing a pre-trained weight $W_0 \in \mathbb{R}^{M \times N}$ and adding a trainable low-rank update $\Delta W = BA$, where $A \in \mathbb{R}^{r \times N}$ and $B \in \mathbb{R}^{M \times r}$. 
During training, the layer output is
\begin{equation}
    h \;=\; W_{0}x + \Delta Wx = W_{0}x + BAx,
    \label{eq:LoRA}
\end{equation}
where $x \in \mathbb{R}^{N}$ is the input and $h \in \mathbb{R}^{M}$ is the output. 
After training, the merged weight is $W_{\text{merged}} = W_0 + \Delta W$, and the auxiliary path is removed, keeping inference-time FLOPs and memory identical to the frozen backbone~\cite{Hu:ICLR22}.

Compared to full fine-tuning, which updates $\#(W_0)=MN$ parameters, LoRA trains only $\#(\text{LoRA}) = r(M+N)$. 
A common initialization sets $B{=}0$ and samples $A$ from a zero-mean Gaussian so that $\Delta W{=}0$ at the start. 
This prevents abrupt output shifts and ensures the same initial output as the pre-trained backbone, but also makes early-stage training sensitive to hyperparameters (e.g., learning rate, variance)~\cite{Hayou:NIPS24, Shiwei:ICML25} and often slows initial accuracy gains~\cite{Meng:NIPS24}. 
Despite these drawbacks, LoRA is widely adopted due to its strong accuracy–efficiency trade-off.

\paragraph*{{Extending LoRA to CNNs}}
LoRA’s matrix assumption aligns naturally with Transformer-style linear layers {but not with CNN convolutional layers, in which weights are multi-dimensional tensors}.
Here, $C_{\mathrm{out}}$ and $C_{\mathrm{in}}$ denote the numbers of output and input channels, respectively, and $k$ denotes the kernel size {(assuming 2D convolutional layers with square kernels $k \times k$ for simplicity).}
To bridge this gap, LoRA-C flattens a convolutional {weight tensor} into a matrix and applies LoRA in the reshaped space~\cite{Ding:arXiv24}.
Concretely, for $W_{\mathrm{conv}} \in \mathbb{R}^{C_{\mathrm{out}} \times C_{\mathrm{in}} \times k \times k}$, LoRA-C introduces
\[
A \in \mathbb{R}^{r \times (k C_{\mathrm{in}})},\quad
B \in \mathbb{R}^{(k C_{\mathrm{out}}) \times r},\quad
\Delta W \in \mathbb{R}^{(k C_{\mathrm{out}})\times(k C_{\mathrm{in}})},
\]
then reshapes $\Delta W$ back to $\mathbb{R}^{C_{\mathrm{out}} \times C_{\mathrm{in}} \times k \times k}$ for merging.

Empirically, LoRA-C often sets the effective rank proportional to the kernel size; 
for square kernels with $r' = k\,r$, the trainable parameter count becomes
\[
\#(\text{LoRA-C}) \;=\; r' \big(k C_{\mathrm{out}} + k C_{\mathrm{in}}\big) 
\;=\; r\,k^2 \,(C_{\mathrm{out}} + C_{\mathrm{in}}),
\]
which grows \emph{quadratically} with the kernel size. 
As a result, LoRA-C extends LoRA to CNNs, but its quadratic parameter growth leads to substantial training-time memory and compute overhead—limiting practicality for on-device learning.

\begin{figure}[t]
 \vskip -6pt
    \centering
    \includegraphics[width=\columnwidth]{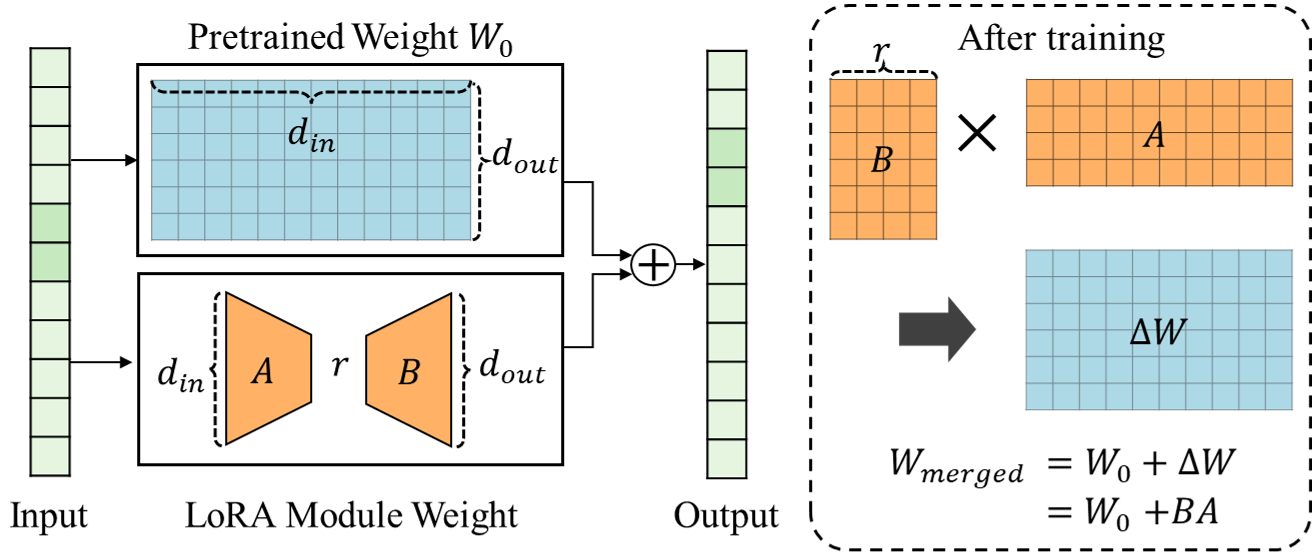}
    \caption{\small LoRA overview with trainable low-rank matrices $A,B$ and the merge into $W_0$.}
    \label{fig:LoRA}
    \vskip -6pt
\end{figure}

\subsection{TTD and TT-SVD for Efficient Adaptation}

\paragraph{TT preliminaries}
Tensor Train Decomposition (TTD) factorizes a $d$-way tensor $\mathcal{W} \in \mathbb{R}^{n_1 \times \cdots \times n_d}$ into cores $\{\mathcal{G}^{(1)}, \ldots, \mathcal{G}^{(d)}\}$ with
\[
\mathcal{G}^{(k)} \in \mathbb{R}^{r_{k-1} \times n_k \times r_k}, \qquad r_0 = r_d = 1,
\]
where TT-ranks $\{r_k\}$ determine both compression and approximation accuracy. 
The parameter count reduces from $\prod_{i=1}^d n_i$ to $\sum_{k=1}^{d} r_{k-1} n_k r_k$, and sequential mode-1 contractions reconstruct a tensor with the original order (approximately, when truncated).

\paragraph{TT-SVD and truncation}
TT-SVD decomposes a multi-dimensional tensor through a sequence of SVDs with rank truncation~\cite{Oseledets:SIAM11}.
Each step reshapes the current tensor view into a matrix, {performs an SVD in the form $U \Sigma V^\top$}, truncates to a target rank, and reshapes $U$ into a core while passing $\Sigma V^\top$ to the next step.
For stability, at TT-SVD step $k$ we set
\[
r_{\text{trunc}} \;=\; \min(r_T, n_k),
\]
where $r_T$ is the target TT-rank and $n_k$ is the maximum attainable rank of the step-$k$ unfolding. 
Repeating this from $k{=}1$ to $d{-}1$ yields $\{\mathcal{G}^{(1)}, \ldots, \mathcal{G}^{(d-1)}\}$, and at the final step ($k{=}d$) the residual $\Sigma V^\top$ is reshaped into the last core $\mathcal{G}^{(d)}$. 
Sequential contractions of all cores reconstruct a tensor with the original dimensions and values up to truncation error.

\paragraph{TT-based PEFT in prior work}
Recent PEFT studies use tensor trains to alleviate matrix-based LoRA limitations by inserting TT adapters into linear layers of LLMs~\cite{Yang:naacl24, Anjum:ICMLA24}. 
Such designs initialize TT cores randomly, discarding information encoded in pre-trained weights and requiring learning from scratch—slowing convergence, which is critical for frequent on-device updates. 
By contrast, SVD-based initialization in matrix LoRA has been shown to improve early-stage convergence by aligning gradients with dominant weight directions~\cite{Meng:NIPS24}. 
TT-SVD extends this rationale to tensors by directly initializing cores from the pre-trained parameter tensor.

\paragraph{Observation: gradient-rank bottleneck in multi-core adapters}
With small TT-ranks, successive contractions can attenuate the effective rank of gradients before they reach inner or input-side cores, analogous to the information bottleneck observed when training only the input-side matrix in LoRA~\cite{Zhu:ICML24}. 
This effect partially explains why output-side updates (nearest to $C_{\mathrm{out}}$) tend to offer stronger early adaptation for CNN convolutions, while keeping the number of trainables small. 
Another practical consideration is that TT-SVD cores closely approximate the pre-trained weights; if all cores are left active at initialization, the layer’s output can be nearly doubled. 
This motivates zero-initializing the first core to keep the auxiliary path inactive until training begins.

\paragraph{Takeaways for CNN PEFT}
(i) Flatten-and-adapt approaches (e.g., LoRA-C) incur rank coupling to the kernel area and quadratic parameter growth in $k$; (ii) TT adapters capture multiway structure but benefit substantially from \emph{pre-trained} initialization (TT-SVD) to avoid cold starts; and (iii) when TT-ranks are small (as required on edge devices), focusing updates near the output side helps preserve gradient information while containing compute and memory during training. 
These insights motivate the structure-preserving, rank-controlled, and selectively trained adaptation pathway developed in our method.

\section{Proposed LoRA-Edge Method}

\begin{figure}
\vskip -10pt
    \centering
    \includegraphics[width=0.97\columnwidth]{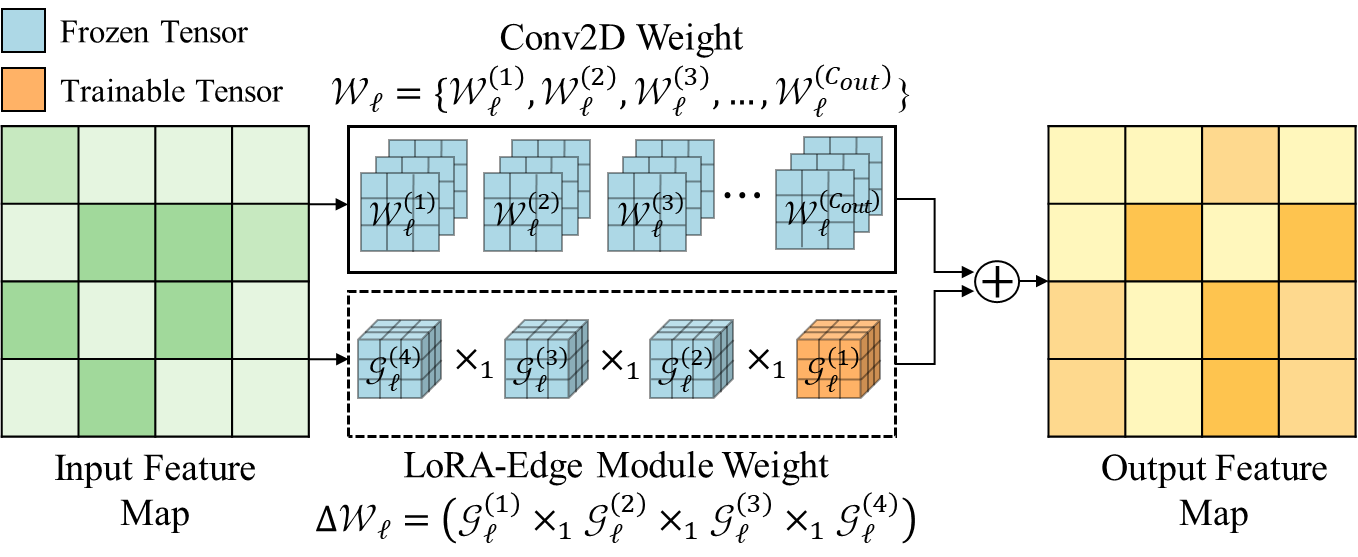}
    \vskip -2pt
    \caption{\small Architectural view of LoRA-Edge.}
    \label{fig:LoRA_EDGE}
    \vskip -6pt
\end{figure}

\subsection{Overview}
As discussed in Sec.~2, matrix-based LoRA couples the adapter rank to the kernel size in CNNs, inflating trainable parameters, while TT-based adapters commonly rely on random initialization, slowing early convergence. 
To address both limitations, we propose {LoRA-Edge}, which integrates {structure-preserving} TT-SVD factorization of pre-trained convolutional layer with {selective} core training.

\refFigure{fig:LoRA_EDGE} illustrates LoRA-Edge on a {2D convolutional layer (Conv2D):} a LoRA-Edge module $\Delta \mathcal{W}$ is connected in parallel to the frozen 4D weight tensor $\mathcal{W}_0$.
During the forward pass, the layer output is the sum of the two convolution paths—one through $\mathcal{W}_0$ and the other through $\Delta \mathcal{W}$. 
Among the TT cores of $\Delta \mathcal{W}$, we freeze \emph{all} except the output-nearest core and \emph{update only} $\mathcal{G}^{(1)}$. 
Since TT-SVD cores closely approximate $\mathcal{W}_0$, directly enabling the parallel path would nearly double the initial layer contribution; we therefore zero-initialize $\mathcal{G}^{(1)}$ so the auxiliary path is inactive at the start. 
This design minimizes trainable parameters while preserving the merge-after-training property.

\begin{algorithm}[!t]
\footnotesize
\caption{LoRA-Edge Fine-Tuning}
\begin{algorithmic}[1] \label{alg:finetuning}
\REQUIRE Pre-trained CNN $\mathcal{M}$, streaming sensor dataset $D_{\mathrm{SS}}$
\ENSURE Adapted CNN $\mathcal{M}_{\mathrm{adap}}$
\STATE \textbf{procedure} \textsc{LoRA\_EDGE\_FINE\_TUNE}($\mathcal{M}$)
\STATE \quad \textbf{Construct LoRA-Edge modules with TT-SVD:}
\STATE \quad \textbf{for} each convolutional layer $\ell$ in $\mathcal{M}$ \textbf{do}
\STATE \qquad Apply TT-SVD to $\mathcal{W}_{\ell}$ with target TT-rank $r_T$; at each SVD step $k$, set $r_{\mathrm{trunc}} \leftarrow \min(r_T, n_k)$.
\STATE \qquad Obtain cores $\{\mathcal{G}_\ell^{(1)}, \ldots, \mathcal{G}_\ell^{(d)}\}$ with $r_0=r_d=1$.
\STATE \qquad Zero-initialize $\mathcal{G}_\ell^{(1)}$; mark it trainable; freeze $\mathcal{G}_\ell^{(2)},\ldots,\mathcal{G}_\ell^{(d)}$ and $\mathcal{W}_\ell$.
\vskip -8pt
\STATE \qquad Define $\Delta \mathcal{W}_\ell \leftarrow \{\mathcal{G}_\ell^{(1)}, \ldots, \mathcal{G}_\ell^{(d)}\}$ and attach in parallel to $\mathcal{W}_\ell$.
\STATE \quad \textbf{end for}
\STATE \quad \textbf{On-device fine-tuning under domain shift:}
\STATE \quad \textbf{for} \textit{step} $= 1$ \textbf{to} \textit{step\textsubscript{max}} \textbf{do}
\STATE \qquad Sample a mini-batch $X$ from $D_{\mathrm{SS}}$; compute $\hat{Y}\mkern-2mu =\mkern-2mu \mathcal{M}.\textsc{forward}(X)$.
\vskip -8pt
\STATE \qquad Evaluate loss $L = \mathcal{L}(Y, \hat{Y})$.
\STATE \qquad Backpropagate gradients \emph{only} to $\mathcal{G}^{(1)}_\ell$ for all layers $\ell$; update $\mathcal{G}^{(1)}_\ell$.
\STATE \quad \textbf{end for}
\STATE \quad \textbf{Merge LoRA-Edge modules into the backbone:}
\STATE \quad \textbf{for} each convolutional layer $\ell$ \textbf{do}
\STATE \qquad  Reconstruct $\Delta\mathcal{W}_\ell$ by mode-1 contractions and set $\mathcal{W}_{\ell} \mkern-1mu \leftarrow \mkern-1mu \mathcal{W}_{\ell} \mkern-1mu + \mkern-1mu\Delta \mathcal{W}_\ell$.
\vskip -8pt
\STATE \quad \textbf{end for}
\STATE \quad \textbf{return} $\mathcal{M}_{\mathrm{adap}} \leftarrow \mathcal{M}$
\STATE \textbf{end procedure}
\end{algorithmic}
\end{algorithm}

\subsection{Detailed Mechanism of LoRA-Edge PEFT}\label{sec:LoRA-Edge_detail}
\refAlgo{alg:finetuning} organizes the process into three phases: (i) \emph{module construction} via TT-SVD, (ii) \emph{on-device fine-tuning}, and (iii) \emph{module merge}.
\begin{table*}[t]
\vskip -6pt
\caption{\small Summary of HAR datasets and cross-domain evaluation protocols. 
Opportunity and DSADS model user variation (LOSO); RealWorld models location variation (LOLO); 
RealDisp models sensor placement variation (ideal vs. self)}
\vskip -3pt
\centering
\renewcommand{\arraystretch}{1.15} 
\resizebox{\textwidth}{!}{%
\small
\begin{tabular}{lccccp{0.35\textwidth}p{0.44\textwidth}}
\Xhline{0.7pt}
Dataset & Subj. & Act. & Freq. & Win. & Sensors & Domain split / Notes \\
\Xhline{0.5pt}
Opportunity~\cite{Chavarriaga:PRL13} 
& 4 & 17 & 30 Hz & 90 
& Wearable, object, ambient; \textbf{wearable only used} 
& By subject (LOSO). Activities include ``drinking from a cup'' and ``opening a door.'' \\

DSADS~\cite{Barshan:TCJ14} 
& 8 & 19 & 25 Hz & 125 
& Body-worn sensors 
& By subject (LOSO). Includes dynamic (``running'', ``cycling'') and static (``sitting'', ``standing'') activities. \\

RealWorld~\cite{Sztyler:PerCom16} 
& 15 & 8 & 50 Hz & 500 
& Seven body locations: chest, forearm, head, thigh, upper arm, waist, shin 
& By location (LOLO). Activities include ``walking'', ``running'', ``climbing stairs''. \\

RealDisp~\cite{Banos:UbiComp12} 
& 17 & 33 & 50 Hz & 250 
& Wearable sensors; \textit{ideal} (predefined) vs. \textit{self} (user-defined) placement; all locations used 
& Split: ideal $\rightarrow$ self. Self placement naturally introduces sensor rotation and displacement. \\
\Xhline{0.7pt}
\end{tabular}
}
\label{tab:datasets}
\vskip -12pt
\end{table*}

\paragraph{Module construction (Lines 2–8)}
For each convolutional layer $\ell$ with weight tensor $\mathcal{W}_\ell$, we apply TT-SVD using a target rank $r_T$ that governs the adaptation capacity and training footprint. 
At each SVD step $k$, the maximum attainable rank $n_k$ can be smaller than $r_T$; thus we use $r_{\mathrm{trunc}}=\min(r_T,n_k)$ during truncation.
TT-SVD yields TT cores $\{\mathcal{G}_\ell^{(1)}, \mathcal{G}_\ell^{(2)}, \mathcal{G}_\ell^{(3)}, \mathcal{G}_\ell^{(4)}\}$ for a
{4D weight tensor}.
Because these cores closely approximate $\mathcal{W}_\ell$, directly enabling the parallel TT path would \emph{double} the layer’s initial contribution. 
To avoid this, we zero-initialize $\mathcal{G}_\ell^{(1)}$ and freeze the remaining cores and $\mathcal{W}_\ell$, making only $\mathcal{G}_\ell^{(1)}$ trainable. 
The resulting $\Delta \mathcal{W}_\ell$ is attached in parallel to the frozen path.

{For 1D convolutional layers (Conv1D), the weight tensor is 3-dimensional, of size $C_{\mathrm{out}} \times C_{\mathrm{in}} \times k$;
this produces TT cores $\{\mathcal{G}^{(1)}, \mathcal{G}^{(2)}, \mathcal{G}^{(3)}\}$.
For clarity, we describe the subsequent mechanism using a 2D convolutional layer}

\paragraph{On-device fine-tuning (Lines 9-14)}
Given streaming data $D_{\mathrm{SS}}$ under domain shift, we iterate for a fixed number of {training} steps. 
Each iteration samples a mini-batch $X$ and performs a forward pass to obtain $\hat{Y}$. 
At layer $\ell$, the output on input feature map $X_\ell$ is
\begin{align}
    Y_{\ell} &= \mathcal{W}_{\ell} * X_{\ell} + \Delta \mathcal{W}_{\ell} * X_{\ell} \nonumber \\ 
      &= \mathcal{W}_{\ell} * X_{\ell}
         + \bigl(\mathcal{G}_{\ell}^{(1)} 
         \times_1 \mathcal{G}_{\ell}^{(2)} \times_1 \mathcal{G}_{\ell}^{(3)} \times_1 \mathcal{G}_{\ell}^{(4)}\bigr) * X_{\ell},
    \label{eq:LoRA_Edge_forward}
\end{align}
where $*$ denotes convolution and $\times_1$ denotes mode-1 contraction along adjacent TT ranks. 
We compute the loss $L=\mathcal{L}(Y,\hat{Y})$ and backpropagate \emph{only} into $\mathcal{G}^{(1)}_\ell$, minimizing gradient compute and parameter updates.

\paragraph{Merge (Lines 15-19)}
After fine-tuning, we reconstruct each $\Delta\mathcal{W}_\ell$ by successive mode-1 contractions,
$\mathcal{G}_\ell^{(1)} \times_1 \mathcal{G}_\ell^{(2)} \times_1 \mathcal{G}_\ell^{(3)} \times_1 \mathcal{G}_\ell^{(4)}$, 
reshape to a 4D weight tensor, and add it to the frozen backbone weight: $\mathcal{W}_\ell \leftarrow \mathcal{W}_\ell + \Delta\mathcal{W}_\ell$. 
The auxiliary path is removed, so inference-time compute and memory equal those of the original model.
After merging across all layers, the original model $\mathcal{M}$ becomes the domain-adapted model $\mathcal{M}_{\mathrm{adap}}$.

\subsection{Theoretical Analysis}

\subsubsection{Parameter Efficiency}
LoRA-Edge achieves high parameter efficiency by combining TT-SVD with selective core training.
For a {2D convolutional weight tensor} $\mathcal{W}\in\mathbb{R}^{C_{\mathrm{out}}\times C_{\mathrm{in}} \times k \times k}$, full fine-tuning updates $k^2 C_{\mathrm{out}}C_{\mathrm{in}}$ parameters.
TT-SVD yields four cores with shapes
\begin{equation}
\begin{array}{rlrl}
\mathcal{G}^{(1)} &\in \mathbb{R}^{1 \times C_{\mathrm{out}} \times r_T}, 
& \quad \mathcal{G}^{(2)} &\in \mathbb{R}^{r_T \times C_{\mathrm{in}} \times r_T}, \\
\mathcal{G}^{(3)} &\in \mathbb{R}^{r_T \times k \times r_T}, 
& \quad \mathcal{G}^{(4)} &\in \mathbb{R}^{r_T \times k \times 1}.
\end{array}
\label{eq:Tensor_Shape}
\end{equation}
We train \emph{only} the first core $\mathcal{G}^{(1)}$, which depends solely on $C_{\mathrm{out}}$ and is therefore unaffected by dynamic input spatial sizes. 
Thus the trainable count reduces from $k^2 C_{\mathrm{out}}C_{\mathrm{in}}$ to $r_T C_{\mathrm{out}}$. 
For example, with $C_{\mathrm{out}}=C_{\mathrm{in}}=64$ and $k=3$, setting $r_T=2$ gives $36{,}864 \rightarrow 128$ ($0.347\%$ of full fine-tuning). 
By contrast, flattening the kernel and applying matrix LoRA trains $r\,k\,(C_{\mathrm{in}}+C_{\mathrm{out}})$ parameters; under the same settings and even with $r=1$, this is $384$, about $3\times$ larger than LoRA-Edge’s $128$. 
In commonly used LoRA-C configurations where the effective rank scales with $k$ (i.e., $r' = k\,r$), the trainables further inflate to $r\,k^2\,(C_{\mathrm{in}}+C_{\mathrm{out}})$, exacerbating the edge training burden.

\subsubsection{Selective Core Training}\label{sec:early_adap}
LoRA-Edge updates only $\mathcal{G}^{(1)}$ based on both computational and optimization considerations.

\textbf{Compute and expressivity.} 
Because $k \ll C_{\mathrm{in}}, C_{\mathrm{out}}$, the spatial-mode cores $\mathcal{G}^{(3)}$ and $\mathcal{G}^{(4)}$ are small and have limited expressivity relative to channel-mode cores {$\mathcal{G}^{(1)}$ and $\mathcal{G}^{(2)}$ (cf. \refEqn{eq:Tensor_Shape}).}
Yet backpropagation through a TT path requires separate gradient contractions for each core; making these small cores trainable adds overhead with modest payoff on edge devices.

\textbf{Gradient information preservation.}
For matrix LoRA, prior work~\cite{Zhu:ICML24} reported that training only the output-side matrix $B$ is more effective than training $A$ alone, while yielding performance nearly comparable to training both $A$ and $B$.

The reason is that during the update of $A$, the gradient $\tfrac{\partial L}{\partial Y}$ is premultiplied by $B^\top \in \mathbb{R}^{r \times d_{\text{out}}}$, reducing its effective rank from the output dimension $d_{\text{out}}$ to $r$ and discarding information.
This gradient rank loss is not unique to matrix LoRA, but also arises in tensor-train adapters.

Specifically, from \refEqn{eq:LoRA_Edge_forward}, the gradients for the input- and output-nearest cores are
\begin{align}
\frac{\partial L}{\partial \mathcal{G}_{\ell}^{(1)}} &=
\frac{\partial L}{\partial Y_{\ell}}\,( X_{\ell}^\top \mathcal{G}_{\ell}^{(4)\top} \mathcal{G}_{\ell}^{(3)\top} \mathcal{G}_{\ell}^{(2)\top}), \nonumber \\
\frac{\partial L}{\partial \mathcal{G}_{\ell}^{(4)}} &=
(\mathcal{G}_{\ell}^{(3)\top} \mathcal{G}_{\ell}^{(2)\top} \mathcal{G}_{\ell}^{(1)\top})\,\frac{\partial L}{\partial Y_{\ell}}\,X_{\ell}^{\top}.
\end{align}
Here, $\tfrac{\partial L}{\partial Y_{\ell}}$ is the gradient with respect to the layer output.
Because TT ranks are intentionally small, repeatedly contracting cores before multiplying by $\tfrac{\partial L}{\partial Y_\ell}$ diminishes its effective rank—analogous to the bottleneck in matrix LoRA.
The same degradation also appears when training only $\mathcal{G}^{(2)}$ or $\mathcal{G}^{(3)}$.

\vspace{2pt}
\textbf{Rationale for training $\mathcal{G}^{(1)}$.}
Selecting the output-side core $\mathcal{G}^{(1)}$ avoids the gradient bottleneck and empirically yields the best adaptation performance.
Meanwhile, the frozen cores $\mathcal{G}^{(2)}, \mathcal{G}^{(3)}, \mathcal{G}^{(4)}$ are initialized by TT-SVD from the pre-trained weight tensor and retain its dominant feature directions~\cite{Meng:NIPS24}. 
As a result, gradients on $\mathcal{G}^{(1)}$ are aligned with the base model and informed by dominant directions retained in the TT-SVD–initialized cores, yielding a more structured starting point for updating $\mathcal{G}^{(1)}$.

We next validate LoRA-Edge through experiments on diverse CNNs and HAR datasets.

\section{Experimental Evaluation}\label{sec:experiments}

\begin{table*}[t]
\vskip -6pt
\caption{F1-score and trainable-parameter ratio across four HAR datasets and three CNN backbones. Results are mean~$\pm$~std for LOSO (Opportunity, DSADS) and LOLO (RealWorld). RealDisp uses a single ideal$\rightarrow$self split (std not applicable).
{Tr. Param. (\%) denotes the fraction of trainable parameters relative to full fine-tuning of the same backbone.}
}
\vskip -2pt
\centering
\renewcommand{\arraystretch}{1.05} 
\resizebox{\linewidth}{!}{
\begin{tabular}{c|c|c|c|cc|cc|cc|cc}
\Xhline{1pt}
\multirow{3}{*}{\textbf{Dataset}} & \textbf{Method} &
\multicolumn{1}{c|}{\textbf{Zero-shot}} &
\multicolumn{1}{c|}{\textbf{Full-FT}} &
\multicolumn{2}{c|}{\textbf{Bias-Tuning}} &
\multicolumn{2}{c|}{\textbf{BN-Tuning}} &
\multicolumn{2}{c|}{\textbf{LoRA-C}} &
\multicolumn{2}{c}{\textbf{LoRA-Edge}} \\
\cline{2-12}
 & \textbf{Model} & \makecell{\textbf{F1-score} \\  \textbf{(\%)}}  &
 \makecell{\textbf{F1-score} \\  \textbf{(\%)}} & 
 \makecell{\textbf{F1-score} \\  \textbf{(\%)}} & \makecell{\textbf{Tr. Param.} \\ \textbf{(\%)}} &
 \makecell{\textbf{F1-score} \\  \textbf{(\%)}} & \makecell{\textbf{Tr. Param.} \\ \textbf{(\%)}} &\makecell{\textbf{F1-score} \\  \textbf{(\%)}} & \makecell{\textbf{Tr. Param.} \\ \textbf{(\%)}} &\makecell{\textbf{F1-score} \\  \textbf{(\%)}} & \makecell{\textbf{Tr. Param.} \\ \textbf{(\%)}} 
 \\
\Xhline{1pt}
\multirow{3}{*}{Opportunity}
 & T-ResNet & 58.3 $\pm$ 1.8 & 90.7 $\pm$ 2.5  & 84.8 $\pm$ 2.3 & 0.49 & 87.2 $\pm$ 2.5 & 0.56 
 & 88.4 $\pm$ 3.8 & 1.10 & 89.9 $\pm$ 2.2 & 0.41 \\
 
 & MobileNet & 63.7 $\pm$ 6.9 & 90.6 $\pm$ 1.5  & 85.7 $\pm$ 2.3 & 1.16 & 86.8 $\pm$ 2.4 & 1.15
 & 87.6 $\pm$ 2.4 & 2.77 & 88.1 $\pm$ 0.9 & 1.15\\
 
 & CALANet   & 52.5 $\pm$ 9.1 & 87.8 $\pm$ 1.6  & 68.9 $\pm$ 4.3 & 0.20 & 70.5 $\pm$ 3.8 & 0.16
 & - & - & 83.1 $\pm$ 0.4 & 0.24\\
\hline
\multirow{3}{*}{DSADS}
 & T-ResNet & 81.4 $\pm$ 8.6  & 99.4 $\pm$ 0.6  & 98.9 $\pm$ 1.2 & 0.51 & 99.2 $\pm$ 1.1 & 0.57
 & 98.5 $\pm$ 1.6 & 1.08 & 99.3 $\pm$ 0.6 & 0.43\\
 
 & MobileNet & 79.5 $\pm$ 10.6  & 99.5 $\pm$ 0.3  & 98.8 $\pm$ 0.5 & 1.32 & 99.3 $\pm$ 0.5 & 1.31
 & 98.8 $\pm$ 0.6 & 2.93 & 99.1 $\pm$ 0.4 & 1.31\\
 
 & CALANet   & 80.3 $\pm$ 8.6  & 99.1 $\pm$ 0.3  & 98.5 $\pm$ 1.3 & 0.21 & 98.3 $\pm$ 1.3 & 0.17
 & -  & - & 99.0 $\pm$ 0.6 & 0.24\\
\hline
\multirow{3}{*}{RealWorld}  
 & T-ResNet & 44.9 $\pm$ 8.7  & 93.7 $\pm$ 2.8  & 84.5 $\pm$ 4.3 & 0.52 & 88.5 $\pm$ 4.0 & 0.58
 & 87.8 $\pm$ 4.1 & 1.06 & 89.5 $\pm$ 0.6 & 0.45\\
 
 & MobileNet & 33.5 $\pm$ 7.8   & 92.1 $\pm$ 2.3  & 28.7 $\pm$ 3.0 & 1.50 & 87.9 $\pm$ 1.9 & 1.49
 & 87.7 $\pm$ 3.1 & 3.17 & 89.8 $\pm$ 2.0 & 1.49\\
 
 & CALANet   & 52.1 $\pm$ 8.4   & 93.5 $\pm$ 4.4  & 86.4 $\pm$ 6.9 & 0.05 & 87.6 $\pm$ 7.2 & 0.05
 & - & - & 91.4 $\pm$ 4.7 & 0.06\\
\hline
\multirow{3}{*}{RealDisp}
 & T-ResNet & 24.89   & 96.2  & 78.9 & 0.48 & 89.6 & 0.55
 & 94.0 & 1.11 & 94.8 & 0.40\\
 
 & MobileNet & 37.28   & 96.1  & 82.7 & 1.08 & 85.2 & 1.07
 & 89.3 & 2.64 & 93.6 & 1.07\\
 
 & CALANet   & 62.59   & 97.8  & 88.6 & 0.09 & 87.5 & 0.08
 & - & - & 95.6 & 0.11\\
\Xhline{1pt}
\end{tabular}}
\label{tab:acc_trainparam}
\vskip -10pt
\end{table*}

We evaluate LoRA-Edge across four HAR datasets and three CNN backbones, analyzing initialization, selective core training, and comparisons against PEFT baselines.

\subsection{Setup and Protocols}\label{sec:TT_LoRA_Exp}

We evaluate LoRA-Edge on four public HAR datasets: Opportunity, DSADS, RealWorld, and RealDisp, under cross–domain settings that reflect typical deployment shifts. 
User variation is captured by Opportunity and DSADS, while location variation is captured by RealWorld and RealDisp.
Dataset characteristics and cross-domain splits are summarized in Table~\ref{tab:datasets}.

In all experiments, training data serve as the \emph{source domain}, and fine-tuning/validation data serve as the \emph{target domain}. For user variation (Opportunity, DSADS), we adopt the Leave-One-Subject-Out (\emph{LOSO}) protocol: one subject is held out as target, and the remaining subjects form the source; this is repeated across all subjects. For location variation (RealWorld), we use Leave-One-Location-Out (\emph{LOLO}): one location is treated as the target while the others form the source; this is repeated across all locations. For RealDisp, data recorded at the \textit{ideal} placement are used as source, and data from the \textit{self} placement—naturally introducing rotation and displacement—are used as target.

To cover both Conv1D- and Conv2D-based HAR models, we use {CALANet~\cite{Park:NIPS24} (Conv1D), MobileNet~\cite{Howard:arXiv17} (Conv2D), and T-ResNet~\cite{Wang:IJCNN17} (Conv2D)}. Since all datasets are imbalanced across classes, we report macro F1-score as the primary evaluation metric.

All methods use Adam~\cite{Kingma:arXiv14}. 
The learning rate is $0.01$ for all methods except \textit{Full fine-tuning} (Full-FT), which uses $0.001$. 
Target-domain data are split $80\%$/$20\%$ into training/test sets. 
Performance is measured at the granularity of {optimization steps}; batch size is fixed to 64. 
When constructing LoRA-Edge via TTD/TT-SVD, we set the TT-rank of each core to $r_T{=}2$.

\begin{figure}[t]
 \vskip -7pt
    \begin{subfigure}{0.495\linewidth}
        \centering
        \includegraphics[width=0.95\textwidth]{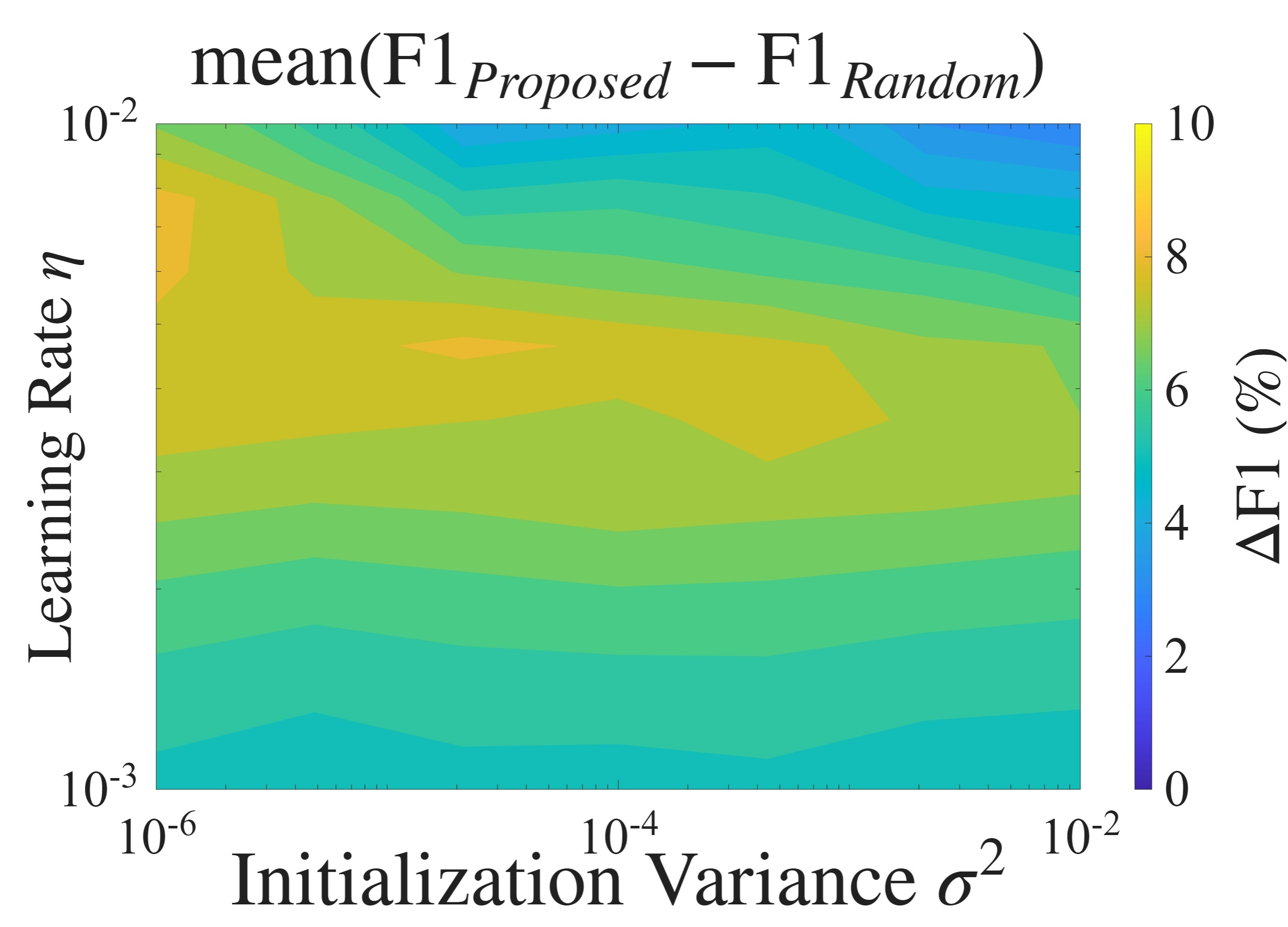}
         \vskip -3pt
        \caption{\footnotesize MobileNet}\label{fig:del_f1_mobilenet}
    \end{subfigure}
    \hfill
    \begin{subfigure}{0.495\linewidth}
        \centering
        \includegraphics[width=0.95\textwidth]{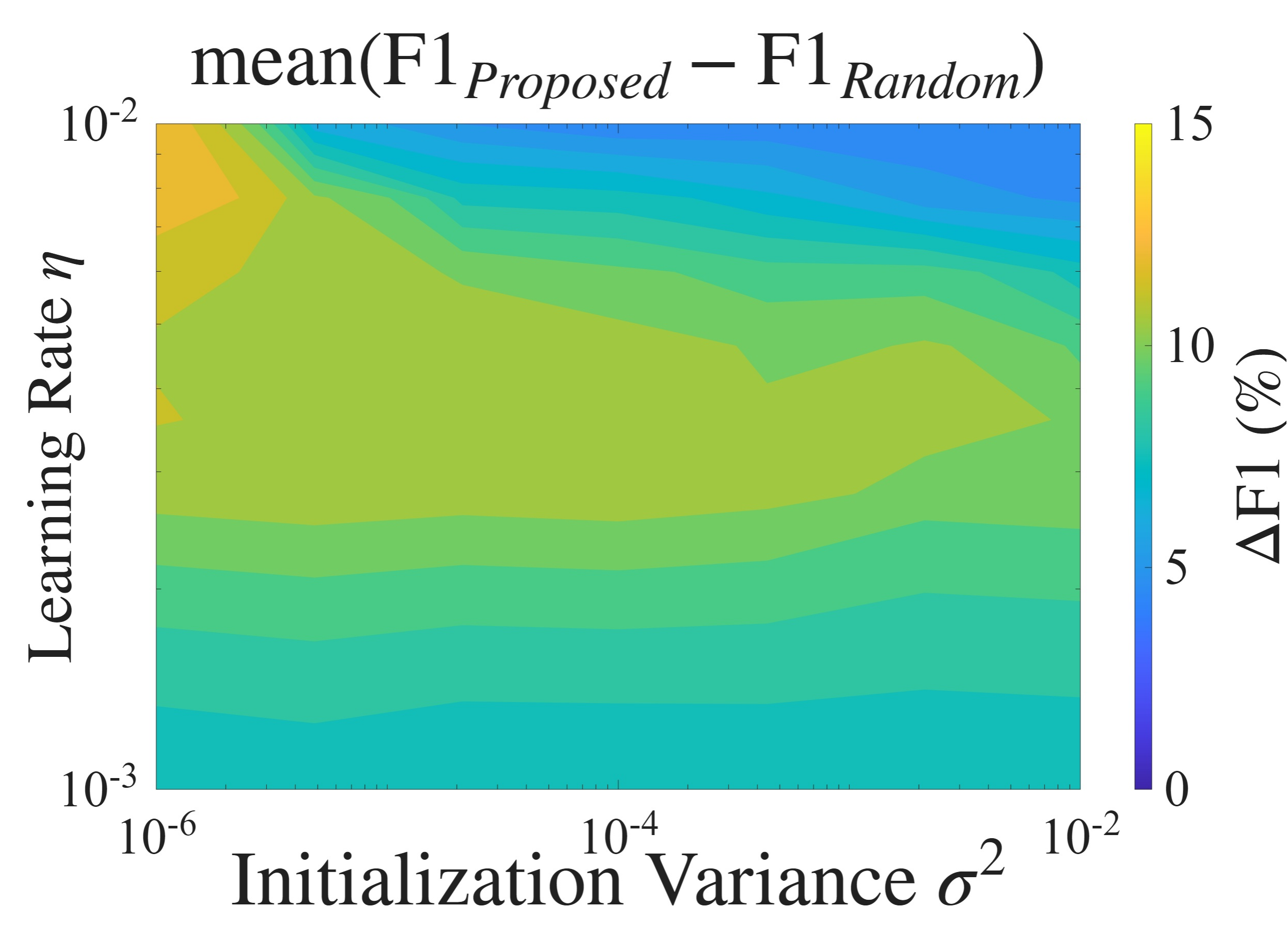}
         \vskip -3pt
        \caption{\footnotesize {T‑ResNet}}\label{fig:del_f1_T-ResNet}
    \end{subfigure}
    \vskip -3pt
     \caption{\small Average F1 difference ($\Delta$F1, \%) between the proposed TT-SVD init. and random init. at the \emph{same} learning rate $\eta$, across $\sigma^2$ {on Opportunity dataset}.}
         \label{fig:del_f1}
     \vskip -6pt
\end{figure}

\subsection{Impact of Initialization}\label{sec:Init_Sensitivity}
Random initialization is sensitive to the variance hyperparameter $\sigma^2$. 
To quantify the effect, we compare the TT-SVD initialization used by LoRA-Edge against random initialization under LOSO on Opportunity, sweeping $\sigma^2$ and learning rate $\eta$. 
Fig.~\ref{fig:del_f1} visualizes the mean F1 difference (proposed minus random) averaged over all held-out subjects.
In the figure, when $\eta$ is small the color variation along the horizontal axis (i.e., across $\sigma^2$) is relatively muted; with larger $\eta$, the horizontal color change becomes pronounced, indicating stronger sensitivity of random initialization to $\sigma^2$. 
Across the entire grid, the average $\Delta$F1 is positive, showing that the proposed initialization achieves higher F1 than random under the same $\eta$ and $\sigma^2$. 
These results indicate that LoRA-Edge’s initialization strategy is {stable} and confers a consistent performance advantage.

\vspace{-4pt}
\subsection{Selective Core Training Analysis}\label{sec:Core_Selection}
We empirically verify the benefits of selective core training and the initialization strategy of LoRA-Edge. 
Specifically, we evaluate six variants: 
\mbox{$\mathcal{G}^{(1)}$ only} (the LoRA-Edge default), 
\mbox{$\mathcal{G}^{(2)}$ only}, 
\mbox{$\mathcal{G}^{(3)}$ only}, 
\mbox{$\mathcal{G}^{(4)}$ only}, 
\textit{All} (train all TT cores initialized by TT-SVD), and 
\textit{All \& zero-init $\mathcal{G}^{(1)}$} (zero-initialize $\mathcal{G}^{(1)}$ and then train all cores). 
We report step-wise F1 over 50 steps on RealWorld with MobileNet and T-ResNet.

\begin{figure}[t]
\vskip -6pt
    \begin{subfigure}{0.495\linewidth}
        \centering
        \includegraphics[width=0.97\textwidth]{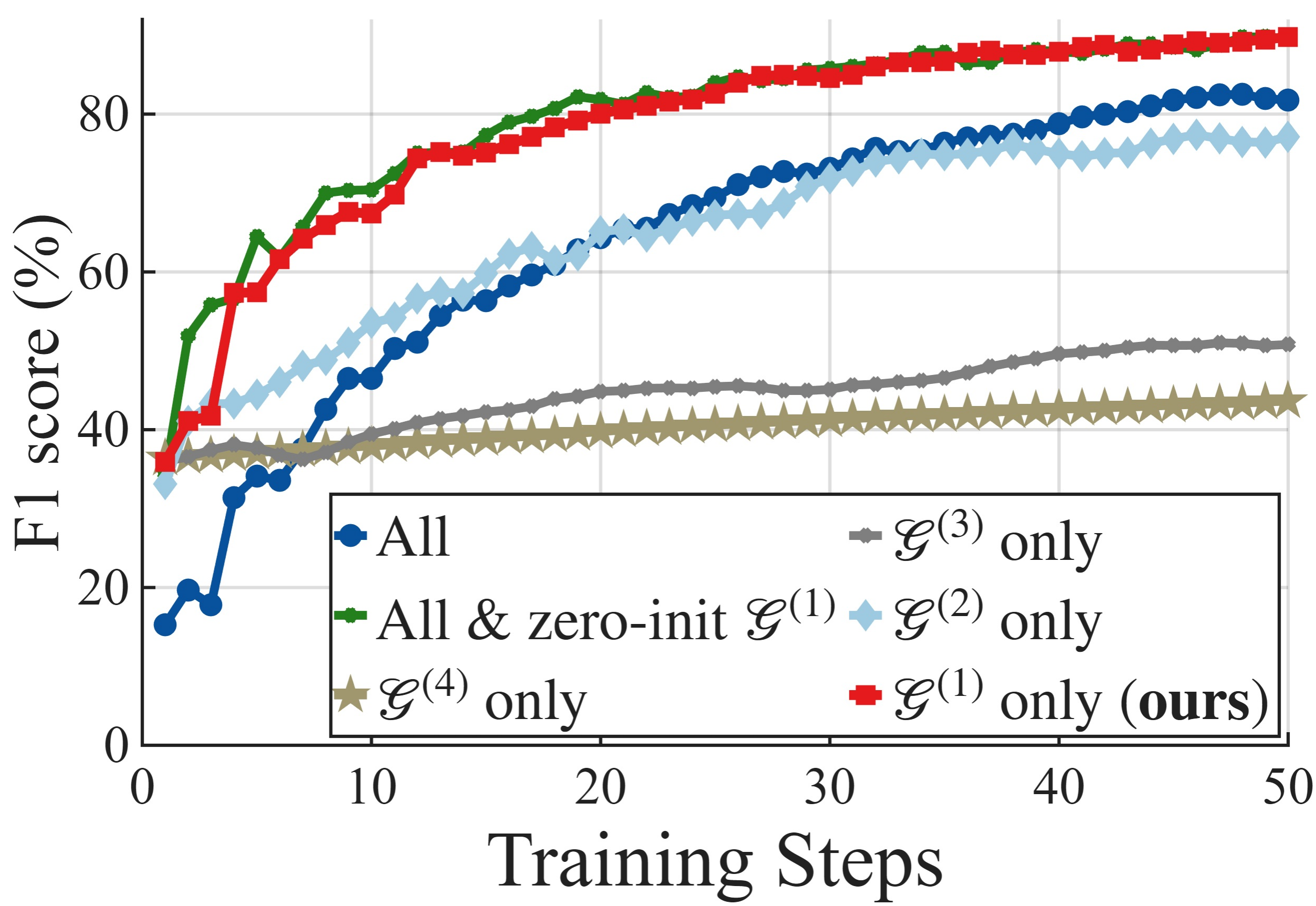}
        \vskip -3pt
        \caption{\footnotesize MobileNet}\label{fig:core_sel_moba}
    \end{subfigure}
    \begin{subfigure}{0.495\linewidth}
        \centering
        \includegraphics[width=0.97\textwidth]{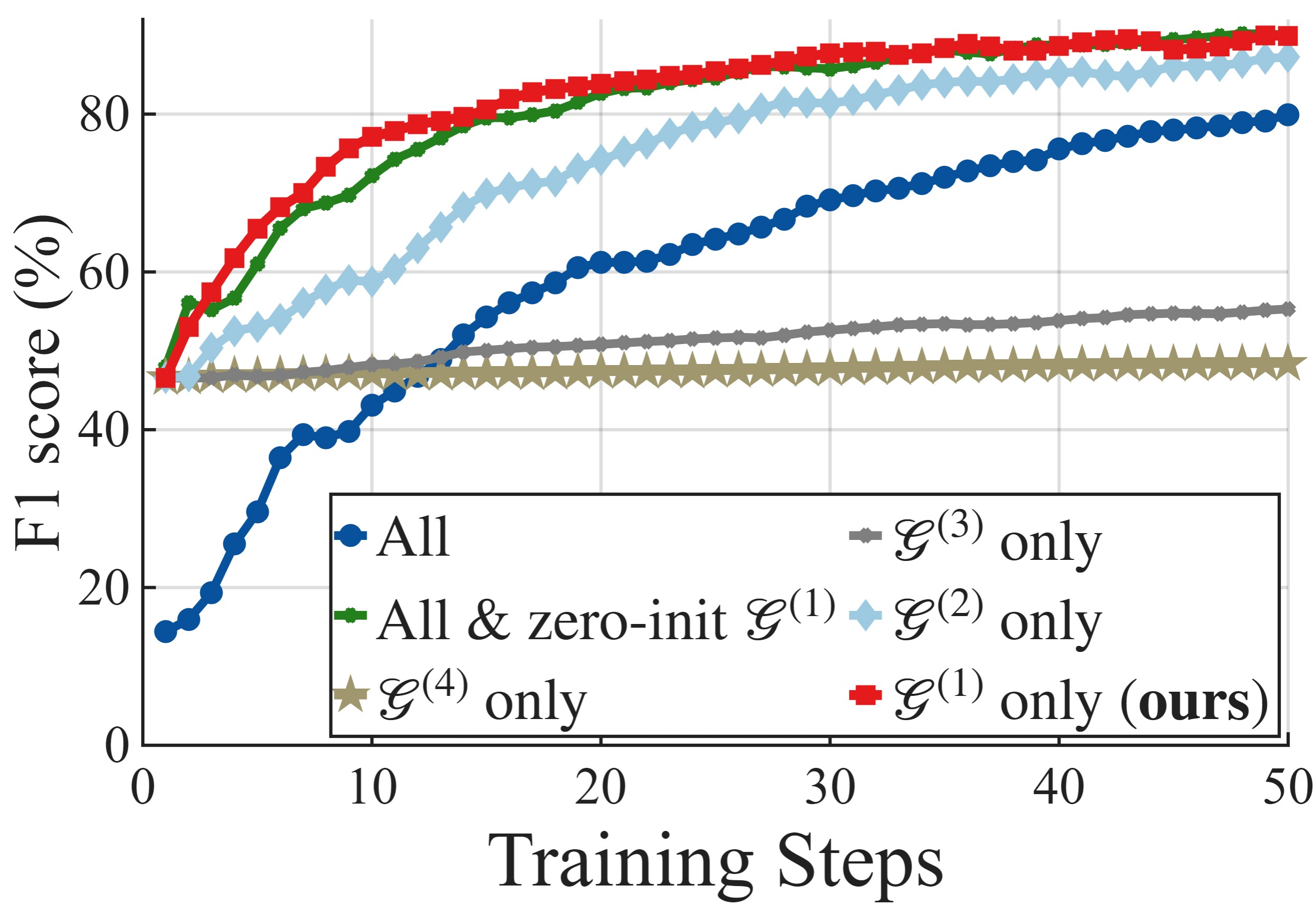}
         \vskip -3pt
        \caption{\footnotesize T-ResNet}\label{fig:core_sel_tres}
    \end{subfigure}
    \vskip -2pt
    \caption{F1 over 50 steps for TT-core training strategies on RealWorld.}\label{fig:core_sel}
   \vskip -6pt
\end{figure}

As shown in \refFigure{fig:core_sel}, \mbox{$\mathcal{G}^{(1)}$ only} (red) and \textit{All \& zero-init $\mathcal{G}^{(1)}$} (green) achieve similarly strong performance and clearly outperform the other variants on both backbones. 
Given that $\mathcal{G}^{(1)}$ constitutes only a small fraction of all TT-core parameters, this demonstrates the parameter-effectiveness of the LoRA-Edge update. 
By contrast, training only the other cores yields weaker gains, which aligns with the limited expressivity and gradient bottlenecks discussed in Sec.~\ref{sec:early_adap}. 
Moreover, the \textit{All} strategy (blue) starts from noticeably lower F1, consistent with the output amplification described in Sec.~\ref{sec:LoRA-Edge_detail} when enabling the TT path without zero-initializing $\mathcal{G}^{(1)}$.

\subsection{Performance Comparison with Other PEFT Methods}\label{sec:Performance_Comparison}
We compare {LoRA-Edge} to {Zero-shot} (source-only), {Full-FT} (full fine-tuning), and CNN-compatible PEFT baselines: {Bias-Tuning}, {BN-Tuning}, and {LoRA-C}. 
Because LoRA-C is designed for Conv2D, it is not applied to CALANet (Conv1D). 
For comparability of trainable budgets, the LoRA-C rank is set to $r{=}1$. 
As a realistic edge constraint, all fine-tuning runs are limited to {50 steps}. 
Note that typical fine-tuning in prior work often uses hundreds of steps~\cite{Honovich:ACL23, Liu:NIPS24, Gupta:NAACL25}, but we restrict to 50 to reflect edge-device constraints.

\refTable{tab:acc_trainparam} reports F1-score and the ratio of trainable parameters. 
Results for Opportunity and DSADS (LOSO) and RealWorld (LOLO) are presented as mean~$\pm$~std across folds, while RealDisp (ideal$\rightarrow$self) is reported as a single F1 value without variance because only one split exists. 
Zero-shot performance corresponds to applying the source-trained model directly to the target domain without fine-tuning.
As expected, {Full-FT} achieves the highest F1 across datasets and models, serving as an upper bound but requiring updates to all parameters—impractical on resource-constrained devices. 
Among PEFT methods, {LoRA-Edge} attains the highest F1 in all cases {except} MobileNet on DSADS, while preserving a small trainable footprint.

Results for Opportunity (LOSO) show that all PEFT methods (except LoRA-C) use $\leq 1.16\%$ trainables. 
LoRA-Edge achieves the highest mean F1 among PEFT methods and shows smaller standard deviations than most alternatives, indicating subject-to-subject consistency.
For DSADS (LOSO), Zero-shot F1 is already relatively high (at least $79.5\%$), reflecting stronger subject similarity. 
All PEFT methods exceed $97\%$ F1 on all backbones, and LoRA-Edge provides competitive means with the smallest or among-the-smallest standard deviations, demonstrating stable adaptation.

In the RealWorld (LOLO) scenario, LoRA-Edge uses $\leq 1.49\%$ trainables across backbones and yields the highest mean F1 for T-ResNet, MobileNet, and CALANet. 
Bias-Tuning produces notably low F1 on MobileNet ($28.7\%$), consistent with its limited capacity when only bias terms are updated~\cite{Cai:arXiv20}. 
While BN-Tuning slightly reduces the standard deviation on MobileNet, LoRA-Edge achieves the lowest standard deviation on T-ResNet and CALANet, underscoring overall consistency across locations.
For RealDisp (ideal$\rightarrow$self), LoRA-Edge uses $\leq 1.07\%$ trainables and attains at least $93.6\%$ F1 on all backbones, outperforming Bias-Tuning (min. $78.9\%$) and BN-Tuning (min. $85.2\%$) at similar parameter ratios. 
This indicates that LoRA-Edge narrows the gap to Full-FT while retaining parameter efficiency under sensor displacement and rotation.

\begin{figure}[t]
\vskip -4pt
\centering
\begin{subfigure}{0.47\columnwidth}
    \includegraphics[width=\textwidth]{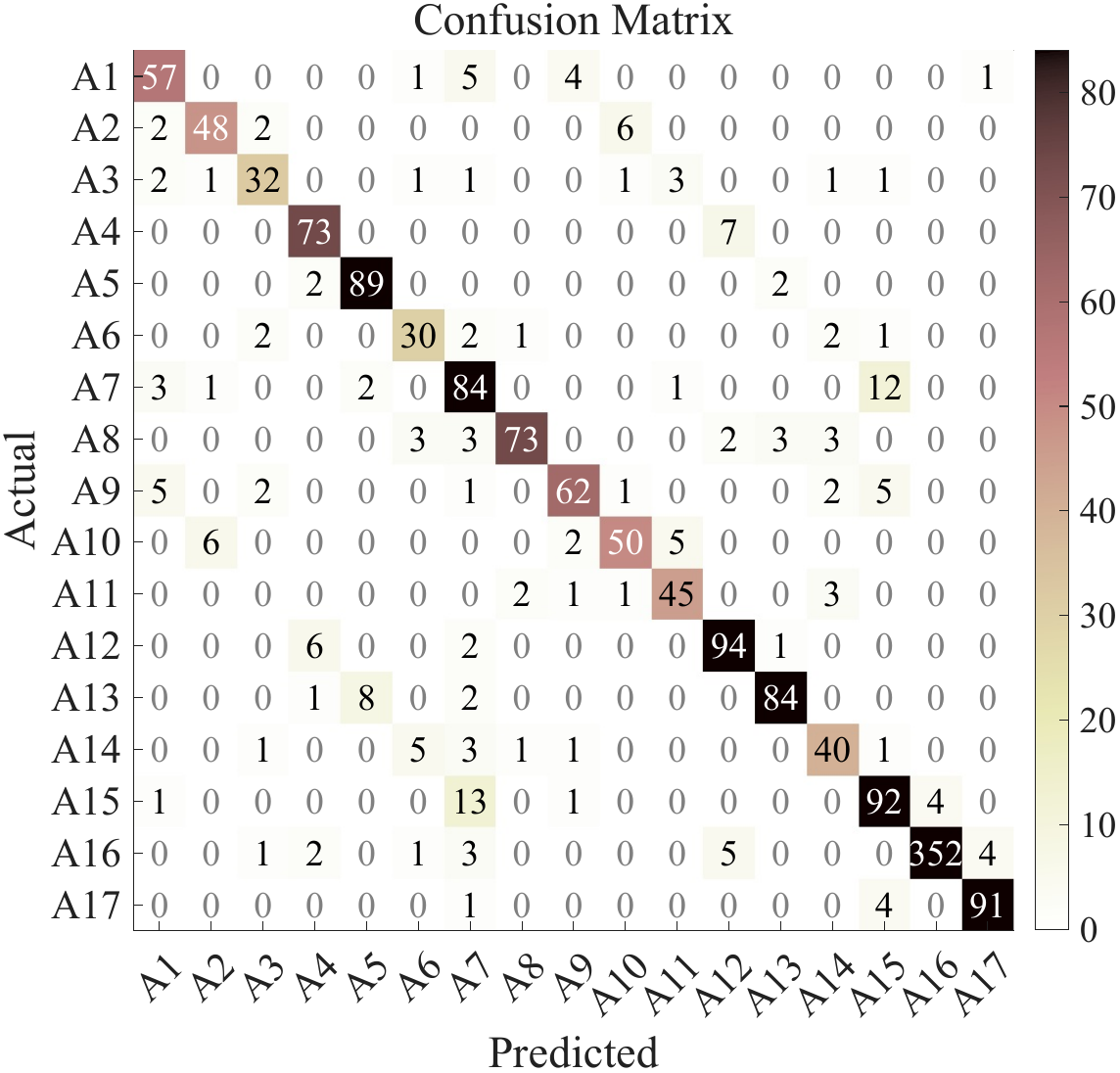}
    \vskip -2pt
    \caption{\footnotesize MobileNet.}
    \label{fig:ConfuseMatrix_moba}
\end{subfigure}
\hfill
\begin{subfigure}{0.47\columnwidth}
    \includegraphics[width=\textwidth]{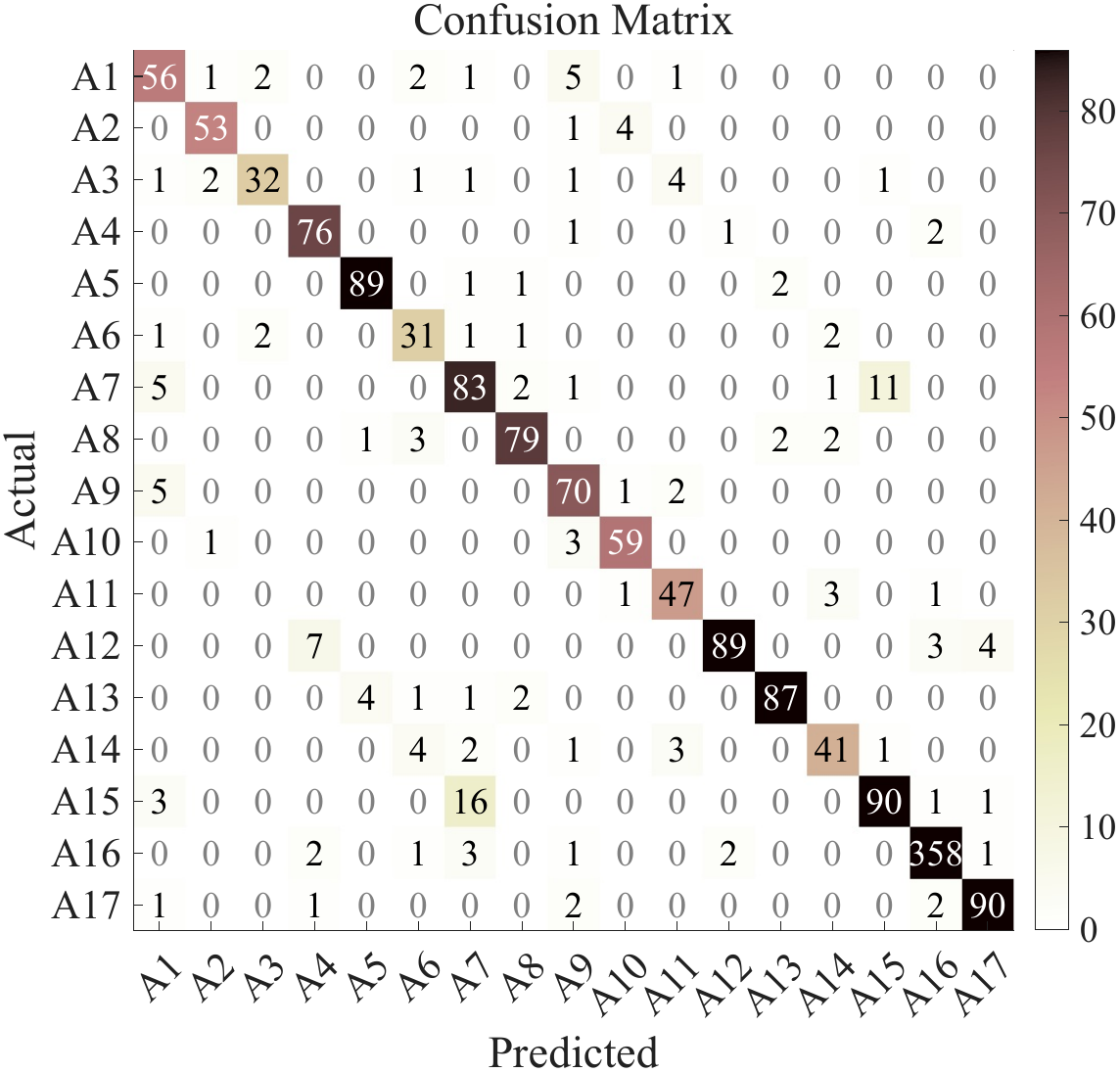}
    \vskip -2pt
    \caption{\footnotesize T-ResNet.}
    \label{fig:ConfuseMatrix_res}
\end{subfigure}
\vskip -3pt
\caption{Confusion matrices of LoRA-Edge on Opportunity.}
\label{fig:ConfuseMatrix}
\vskip -6pt
\end{figure}

\subsection{Error Analysis with Confusion Matrices}
\refFigure{fig:ConfuseMatrix} shows confusion matrices on the Opportunity dataset (A1–A17) when applying LoRA-Edge with MobileNet and T-ResNet. 
The x-axis denotes predicted labels and the y-axis denotes ground truth. 
Per-subject matrices are computed and aggregated across subjects. 
Both models exhibit strong diagonal structure, indicating stable classification performance. 
Notably, A5 (\textit{Open Fridge}) and A16 (\textit{Drink from Cup}) exceed $95\%$ accuracy. 
A pronounced mutual confusion is observed between A7 (\textit{Open Dishwasher}) and A15 (\textit{Clean Table}), indicating that certain activity pairs are inherently difficult to separate. 
Nevertheless, LoRA-Edge overall demonstrates effective class discrimination across the 17 activities.

\subsection{Convergence Speed on Edge Devices}
To demonstrate the practicality of LoRA-Edge on real edge devices, we compare the convergence time of different PEFT methods using an NVIDIA Jetson Orin Nano development kit~\cite{Jetson}. 
The Jetson Orin Nano is equipped with an Ampere GPU with 1,024 CUDA cores and 8\,GB of memory, providing a realistic edge-AI computing environment for HAR applications.

For clarity of comparison, convergence-time analysis is reported on the RealDisp dataset. 
Unlike other datasets (e.g., Opportunity, DSADS, RealWorld), which involve many possible target domains depending on subject or location splits, RealDisp in our setup provides a single ideal$\rightarrow$self partition, avoiding confounding effects from target-domain selection. 
A broader evaluation across multi-domain settings is left outside the scope of this analysis.

\begin{figure}[t]
\vskip -4pt
\centering
\begin{subfigure}{\linewidth}
    \centering
    \includegraphics[width=\linewidth]{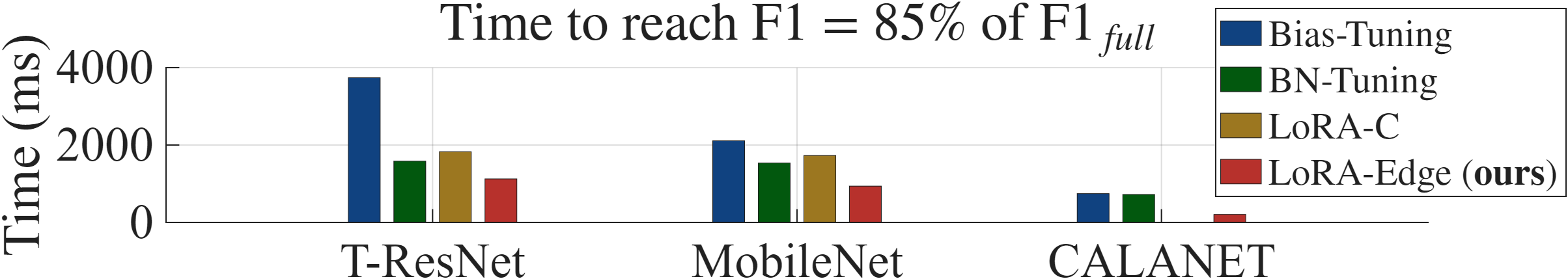}
     \vskip -2pt
    \caption{\footnotesize 85\% of Full-FT}
    \label{fig:Time_to_TargetF1_85}
\end{subfigure}
\vskip 2pt
\begin{subfigure}{\linewidth}
    \centering
    \includegraphics[width=\linewidth]{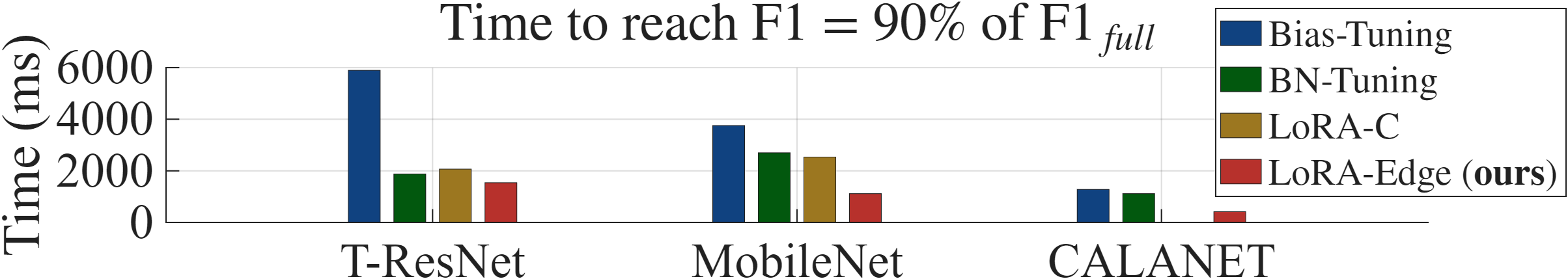}
     \vskip -2pt
    \caption{\footnotesize 90\% of Full-FT}
    \label{fig:Time_to_TargetF1_90}
\end{subfigure}
 \vskip -3pt
\caption{Convergence time of PEFT methods on the Jetson Orin Nano at 85\% and 90\% of Full-FT.}
\label{fig:Time_to_TargetF1}
\vskip -6pt
\end{figure}

Using the Full-FT F1 in \refTable{tab:acc_trainparam} as the upper bound, we measure the time required for each PEFT method to reach {85\%} and {90\%} of this bound. 
\refFigure{fig:Time_to_TargetF1_85} and \refFigure{fig:Time_to_TargetF1_90} show the convergence times on RealDisp with three CNN backbones: T-ResNet, MobileNet, and CALANet. 
We compare against Bias-Tuning, BN-Tuning, and LoRA-C; note that LoRA-C was originally designed for Conv2D layers and is therefore excluded from experiments with CALANet, which is based on Conv1D.

\refFigure{fig:Time_to_TargetF1_85} presents the time to reach 85\% of the Full-FT upper bound. 
Across all CNN backbones, LoRA-Edge converges the fastest: on average, it achieves the target performance within one second, whereas the other PEFT baselines require approximately $1.4\times$ to $3.5\times$ longer. 
This advantage is consistently observed at the 90\% threshold in \refFigure{fig:Time_to_TargetF1_90}. 
In particular, Bias-Tuning requires about $3.8\times$ more time than LoRA-Edge to reach 90\%, highlighting its limited practicality when higher accuracy is required.

Through this convergence-time comparison on a representative off‑the‑shelf edge-AI platform, we verify that LoRA-Edge converges \emph{faster and more stably} than previously proposed PEFT methods. 
These results demonstrate that LoRA-Edge is a feasible on-device PEFT approach capable of achieving effective real-time domain adaptation even under strict computational and power constraints.

\section{Conclusion}  
  
We presented \textit{LoRA-Edge}, a structure-preserving PEFT framework for CNNs that applies TT-SVD to pre-trained convolutional layers, selectively updates only the output-side core with zero-initialization, and merges the update post-training to preserve inference efficiency.  
Across four HAR datasets and three CNN backbones, LoRA-Edge achieved accuracy within {4.7\%} of full fine-tuning while training at most {1.49\%} of parameters. On-device evaluation further confirmed its practicality: LoRA-Edge converged up to {3.8}$\times$ faster on a Jetson Orin Nano, demonstrating both training efficiency and deployment feasibility under strict edge budgets.  
These results highlight that merge-and-run adaptation, aligned with the tensor structure of CNN weights, turns parameter-efficient fine-tuning into a practical mechanism for real-world on-device learning.   



\bibliographystyle{IEEEtran}
\bibliography{reference}
\end{document}